 \documentclass[final,5p,times]{elsarticle}

\usepackage{amssymb}
\usepackage{amsmath}
\usepackage{hyperref}
\usepackage{xcolor}

\journal{Computer Methods and Programs in Biomedicine}

\begin{document}

\begin{frontmatter}



\title{Analysis and Evaluation of Handwriting in Patients with Parkinson's Disease Using kinematic, Geometrical, and Non-linear Features}
 

\author{C. D. Rios-Urrego$^{1}$, J. C. V\'asquez-Correa$^{1,2*}$\corref{cor1}, J. F. Vargas-Bonilla$^{1}$, E. N\"oth$^{2}$, F. Lopera$^{3}$, and J. R. Orozco-Arroyave$^{1,2}$}

\address{$^{1}$Faculty of Engineering, University of Antioquia UdeA, Medell\'in, Colombia.\\
$^{2}$Pattern Recognition Lab, Friedrich-Alexander-Universit{\"a}t Erlangen-N{\"u}rnberg, Germany.\\
$^{3}$Neuroscience Research Group, Faculty of Medicine, University of Antioquia UdeA, Medell\'in, Colombia.}
\cortext[cor1]{corresponding author J. C. V\'asquez-Correa\\ e-mail: juan.vasquez@fau.de\\ Address: Friedrich-Alexander-Universität Erlangen-Nürnberg (FAU)
Lehrstuhl f\"ur Informatik 5 (Mustererkennung)
Martensstr. 3, Of 10.158,
91058 Erlangen
Germany}

\begin{abstract}
\textbf{Background and objectives:} Parkinson's disease is a neurological disorder that affects the motor system producing lack of coordination, resting tremor, and rigidity. 
Impairments in handwriting are among the main symptoms of the disease. 
Handwriting analysis can help in supporting the diagnosis and in monitoring the progress of the disease. This paper aims to evaluate the importance of different groups of features to model handwriting deficits that appear due to Parkinson's disease; and how those features are able to discriminate between Parkinson's disease patients and healthy subjects.\\
\textbf{Methods:} Features based on kinematic, geometrical and non-linear dynamics analyses were evaluated to classify Parkinson's disease and healthy subjects. Classifiers based on K-nearest neighbors, support vector machines, and random forest were considered.\\
\textbf{Results:} 
Accuracies of up to $93.1\%$ were obtained in the classification of patients and healthy control 
subjects. 
A relevance analysis of the features indicated that those related to speed, acceleration, and pressure are the most discriminant.
The automatic classification of patients in different stages of the disease shows $\kappa$ indexes
between $0.36$ and $0.44$. Accuracies of up to $83.3\%$ were obtained in a different dataset used only for validation purposes.\\
\textbf{Conclusions:} The results confirmed the negative impact of aging in the classification process when we considered different groups of healthy subjects. In addition, the results reported with the separate validation set comprise a step towards the development of automated tools to support 
the diagnosis process in clinical practice.
\end{abstract}


\end{frontmatter}

\section{Introduction}
\label{sec:introduction}

Parkinson's disease (PD) is a neuro-degenerative disorder, which produces motor and non-motor symptoms in the patients. Examples of motor impairments include resting 
	tremor, bradykinesia, rigidity, micrographia, hypomimia, and others~\cite{hornykiewicz1998}. 
While examples of non--motor impairments include depression, sleep disorders, dementia, 
and others~\cite{PD2004}. 
These symptoms appear as consequence of the progressive loss of dopaminergic 
neurons in the mid-brain~\cite{hornykiewicz1998}. Handwriting is one of the 
most impaired motor activities in PD patients.
The most common symptoms in handwriting of PD patients include micrographia, which is related to the reduction of the size in handwriting, and dysgraphia, which is related to difficulties performing the controlled fine motor movements required to write~\cite{letanneux2014micrographia}.
The progression of motor symptoms is currently evaluated with the third section 
of the Movement Disorder Society -- Unified Parkinson's Disease Rating Scale (MDS-UPDRS-III)~\cite{Goetz2008}, which is administered by neurologist experts.
The scale contains several aspects to evaluate motor skills of the patients, 
including those directly or indirectly related with  
handwriting like finger tapping, hand tremor, hand rigidity, and others.
The diagnosis process of PD is expensive and time-consuming for patients, caregivers, and 
the health system~\cite{pahwa2013handbook,schrag2002valid,johnson2013economic}. 
Automatic handwriting analysis could help to support the process to 
diagnose and monitor the neurological state of the patients.

There is interest in the research community  to automatically assess the
handwriting of PD patients. Most of the studies 
consider data from on-line handwriting, which are captured from digitizer 
tablets, and contain information related to the dynamics of the 
handwriting process while the patient is putting the pen on the tablet 
(i.e., on-surface). Some of the features that are typically extracted from the
on-surface dynamics include the pressure of the pen, and the azimuth and 
altitude angles.
There are also tablets that allow to capture information of the in-air movement 
(i.e., before and/or after the patient to put the pen on the tablet's surface). 
The classical way of addressing the analysis of on-line handwriting is based
on on-surface features.
For instance in~\cite{sarbaz2013separating} the authors proposed 
features related to the power 
spectral density of the speed stroke in the handwriting of 17 healthy 
control (HC) subjects and 13 PD patients. 
The authors classified PD patients vs. HC subjects using a neural network 
and reported an accuracy of up to 86.2\%.
In~\cite{rosenblum2013handwriting} the authors extracted kinematic features from 
the handwriting of 20 HC subjects and 20 PD patients who wrote their addresses and 
full names. The authors computed features
related to the average time on the tablet's surface, the average in-air time of the pen, 
the speed of the trajectory, and the average pressure of the pen. The authors reported
accuracies of up to 97\% in the classification of PD patients vs. HC subjects.
In~\cite{drotar2014analysis} the authors evaluated the in-air movement during the 
handwriting process of 37 PD patients and 38 HC subjects who wrote the Czech 
sentence: \emph{Tramvaj dnes uz nepo-jede} (the train will not go today). 
The authors extracted several kinematic features including the average in-air time 
of the pen, and the average on-surface time. 
The authors performed the classification of PD patients and HC subjects using a 
support vector machine (SVM) with a Gaussian kernel, and reported accuracies 
around 86\% when the features based on pressure, kinematic, in-air time, 
and on-surface time are combined.
In~\cite{drotar2016evaluation} the authors analyzed different handwriting 
tasks performed by the same set of patients considered in \cite{drotar2014analysis}. 
The set of tasks included the Archimedean spiral, the repetition of 
the graphemes \textit{l} and \textit{le}, simple orthographic isolated words, 
and the Czech sentence: \emph{Tramvaj dnes uz nepo-jede}. 
The authors extracted kinematic and pressure-based features to classify PD patients 
and HC subjects. Three different classification methods were considered: 
SVM, K-nearest neighbors (KNN), and Adaboost. 
Accuracies of up to 76.5\% were reported with the combination of kinematic and 
pressure-based features.
%
%
In~\cite{Vasquez2017ICASSP} the authors applied the Generalized Canonical 
Correlation Analysis (GCCA) to classify PD patients and HC subjects, and to predict 
the neurological state of the patients using information from handwriting, speech, 
and gait. 
Handwriting analyses were based on kinematic features extracted from 
the vertical and horizontal axes, and the pressure of the pen. 
The authors concluded that the combination of speech, handwriting, and gait 
using the GCCA approach improves the accuracy in the classification and the prediction 
of the neurological state.

Other neurological disorders have also been studied using online handwriting. 
In~\cite{lopez2016automatic} the authors proposed non-linear dynamics (NLD) 
features to evaluate the handwriting of 10 HC subjects and 25 patients with 
essential tremor. 
Several NLD features such as the fractal dimension and 
Shannon entropy were extracted from the Archimedean spiral.
The NLD features were combined with standard kinematic measures related to 
the speed and pressure of the pen. 
The authors reported accuracies of up to 85\% classifying the patients vs. HC 
subjects.  

Recent deep learning approaches have also been used to classify 
	handwriting of PD and HC subjects. In~\cite{Pereira2018} the authors modeled 
	the handwriting dynamics of 18 HC and 74 PD subjects using a smart-pen with 
	several sensors. The participants performed several tasks including the 
	drawing of circles and Archimedian spirals. The authors proposed a model 
	based on convolutional neural networks (CNNs). The signals from the smart-pen 
	were transformed into images by concatenating and reshaping the time-series. 
	Several CNN configurations were considered for each task, and a majority 
	voting scheme was implemented to make the final decision. 
	Accuracies up to 95\% were reported by the authors.
In \cite{Gallicchio2018} the authors considered a model called deep echo state network 
to classify 67 PD patients and 15 HC subjects who drew Archimedian spirals on a tablet. 
The deep learning model was based on recurrent neural networks (RNNs), which process 
the time-series of horizontal and vertical movements, the grip angle, and the pressure 
of the pen. The authors reported accuracies of up to 88.3\% with the proposed model.
Recently, in~\cite{vasquez2018multimodal} researchers from our research Lab considered 
a deep learning model based on CNNs to classify PD patients and HC subjects using 
multimodal information from speech, handwriting, and gait. 
The handwriting analysis was based on the transition when the patients have the 
pen in the air and put it on the tablet’s surface. The trajectories from the 
pen and the pressure of the pen were modeled with a CNN. The authors reported 
accuracies of up to 97.6\% when information from speech, handwriting, and gait is combined; 
however, the accuracy obtained considering only the handwriting analysis was around 66.5\%.

According to the reviewed literature, most of the studies considered kinematic features 
to evaluate handwriting impairments of patients with neurodegenerative disorders.  There are some recent studies that consider deep learning methods to assess the handwriting deficits of PD patients.
Other kinds of features including geometrical, spectral, and NLD based have not been extensively studied to support the diagnosis and monitoring of the patients. The analysis of these aspects will be the main outcomes of the current study. 

\subsection{Research objectives of this study}

This paper evaluates the performance of three different feature sets 
	(kinematic, geometrical, and NLD) to model handwriting impairments of PD patients. 
	The method aims to provide suitable insights about the relevance of different 
	features to discriminate between PD patients and HC subjects. 
	The proposed approach is evaluated in realistic conditions, i.e., with a different 
	sets of patients and healthy subjects that never participated in the training 
	process. Finally, the study aims to evaluate the suitability of the proposed approach 
	to automatically classify patients in different stages of the disease.

\section{Data Acquisition and Participants}

\begin{figure}[!ht]
	\centering
	\includegraphics[width=0.8\linewidth, trim={0cm, 0cm, 0cm, 2cm}, clip]{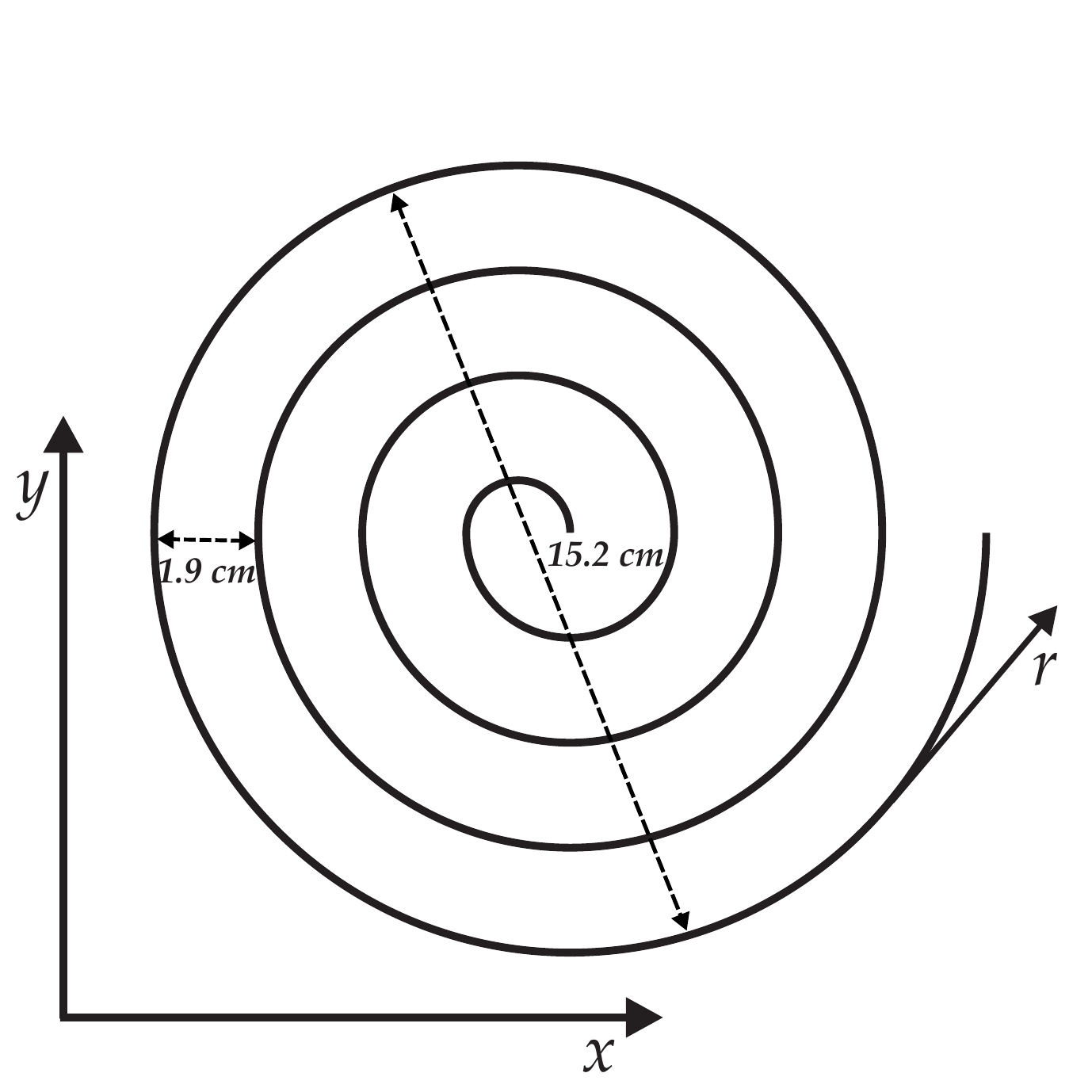}
	\caption{Template of the Archimedean spiral.}
	\label{fig:trajectory}
\end{figure}

The data were collected with a Wacom Cintiq 13 HD tablet, with visual 
feedback to the patients and using a sampling frequency of 180\,Hz. 
The tablet captures six different signals: horizontal position ($x(t)$), vertical 
position ($y(t)$), azimuth angle, altitude angle, distance to the tablet 
surface ($z(t)$), and pressure of the pen.
For this study the participants draw an Archimedean spiral following a predefined
template (Figure~\ref{fig:trajectory}), which was displayed on the 
tablet. Participants were requested to draw the spiral between the lines of the
template and avoiding to cross them. Additionally, the participants wrote the sentence
\emph{``El abecedario es a, b, c,..., z''} which in English is: 
\emph{``The alphabet is a, b, c..., z''}.

\begin{figure*}[!ht]
	\centering
	\includegraphics[width=0.8\linewidth, trim={0cm, 0cm, 0cm, 0cm}, clip]{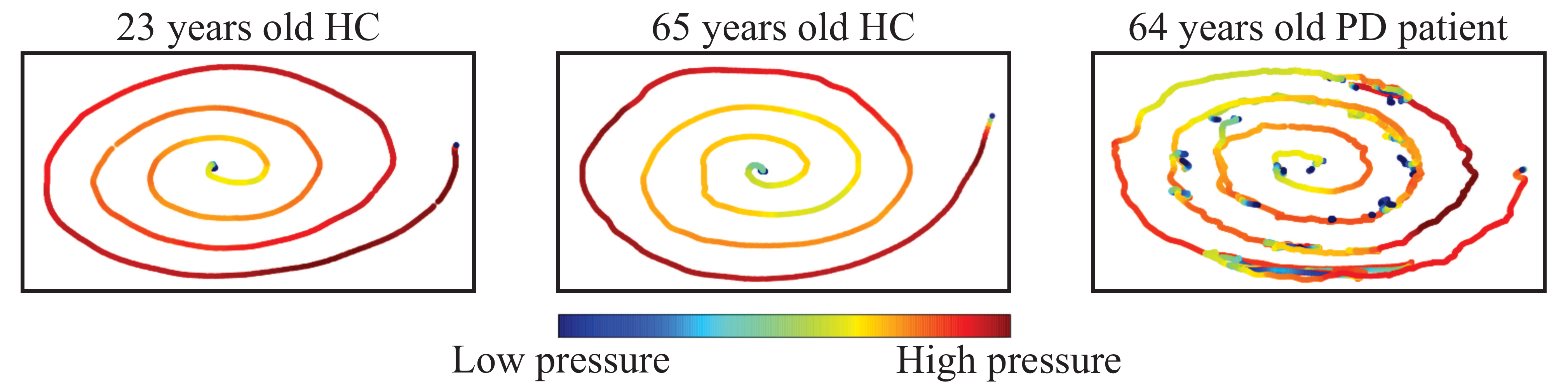}
	\caption{Archimedean spiral drawn by three participants.}
	\label{fig:spiral}
\end{figure*}

\begin{figure*}[!ht]
	\centering
	\includegraphics[width=0.8\linewidth, trim={0cm, 0cm, 0cm, 0cm}, clip]{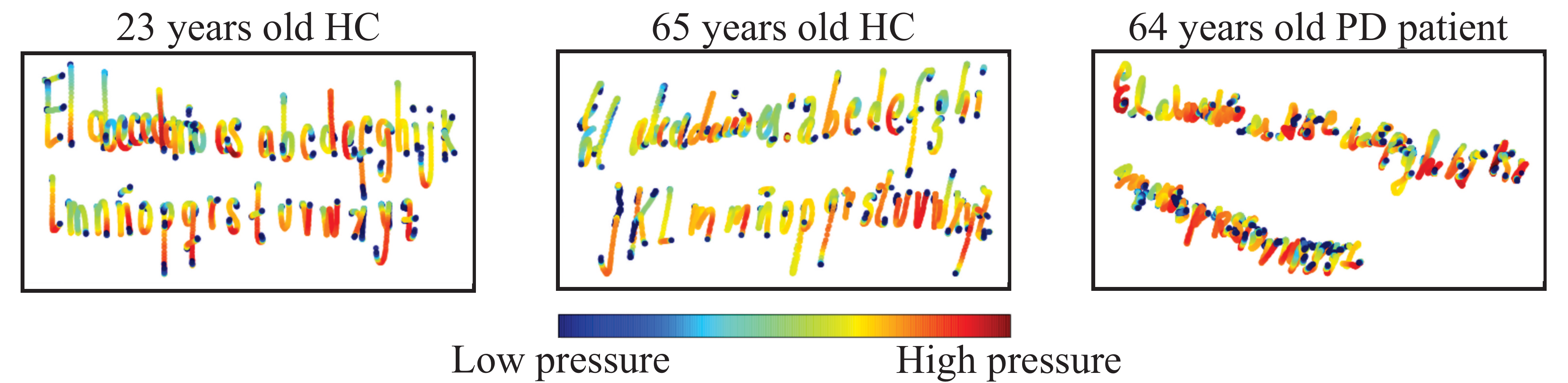}
	\caption{Sentence written by a yHC (left), an eHC (middle), and a PD patient (right). The sentence is: \emph{El abecedario es a b c ... z}.}
	\label{fig:sentence}
\end{figure*}

With the aim to show real cases of how people with different
age and health condition draw, 
Figures \ref{fig:spiral} and \ref{fig:sentence} include examples of drawings
obtained from three different participants: 23 years old subject (left), 65 years old 
subject (middle), and 73 years old PD patient with a MDS-UPDRS-III score of 64 (right).


\subsection{Subjects}

Four groups of participants are considered in this study: 
patients with PD; elderly HC subjects; 
young HC subjects; and an additional set of 12 participants,
which is considered to validate the generalization
capability of the models.
Details of each group are provided below.

\emph{PD patients:}

This group is formed with 39 patients (26 female) with ages between 29 and 81 years
(mean = 59.08; SD = 11.17). 
All of them were evaluated by a neurologist expert according to the 
MDS-UPDRS-III scale~\cite{Goetz2008}. The scores of such evaluations ranged between
8 and 106 (mean = 35.72; SD = 23.25).
This group of patients was recruited in the Foundation for Parkinson's 
patients in Medell\'in, Colombia. The patients were
evaluated during the ``ON'' phase of medication, i.e., no more than three
hours after the intake.

\emph{Elderly HC subjects (eHC):}

This group includes 39 healthy subjects (18 female) with ages between 29 and 85 years 
(mean = 61.85; SD = 13.89). 
None of the participants in this group had symptoms of neurological or movement 
disorders. 
The PD and eHC groups are matched for age ($t(0.99) = -1.19, p = 0.12 $).

\emph{Young HC subjects (yHC):}

A total of 40 healthy participants (16 female) are included in this group. The
age ranges from 20 to 42 years (mean = 24.43; SD = 4.08). Most of the participants of
this set are students who wanted to participate in the study. None of them exhibited or
manifested symptoms of neurological or movement disorders.

\begin{figure}[!ht]
	\centering
	\includegraphics[width=0.8\linewidth, trim={0cm, 0cm, 0cm, 0cm}, clip]{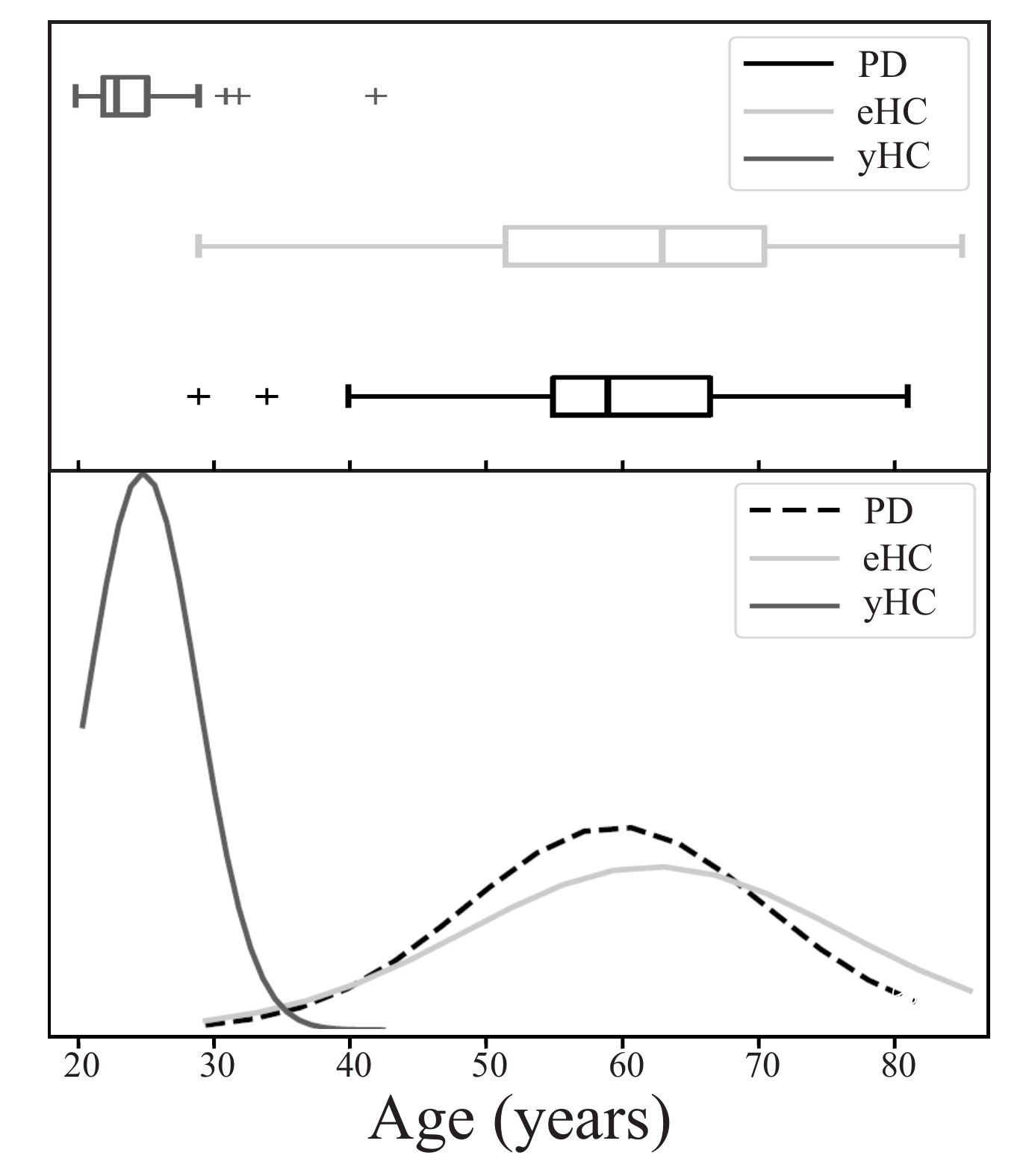}
	\caption{Age distribution of the three groups of subjects.}
	\label{fig:distribution}
\end{figure}

Table~\ref{tab:data} includes demographic and clinical data of the subjects included
in the aforementioned three groups. Figure~\ref{fig:distribution} shows
the age distribution of the three groups of subjects.

\emph{Additional validation set:}

This group of subjects was considered to validate the generalization capability of the
model proposed in this paper.
The group includes 6 PD patients and 6 HC subjects, there are 3 female in each group,
and the ages range from 54 to 83 (mean = 68.33; SD = 10.19) and from 45 to 57 (mean = 53; 
SD = 4.69), respectively.
Demographic and clinical data, including the MDS-UPDRS-III score of each patient in 
this additional validation set, are included in 
Table~\ref{tab:test_data}.

\begin{table*}[!ht]
\caption{Demographic and clinical information of the participants. \textbf{PD}: Parkinson's disease, \textbf{eHC}: elderly healthy controls, \textbf{yHC}: young healthy controls, $\mathrm{\mu}$: average, $\mathrm{\sigma}$: standard deviation.}
\label{tab:data}
\centering

\begin{tabular}{l|cc|cc|cc}
\hline
                                            & \multicolumn{2}{c|}{\textbf{PD patients}}                                & \multicolumn{2}{c|}{\textbf{eHC}}                                 & \multicolumn{2}{c}{\textbf{yHC}}                                   \\
                                            & \multicolumn{1}{c}{\textbf{Male}} & \multicolumn{1}{c|}{\textbf{Female}} & \multicolumn{1}{c}{\textbf{Male}} & \multicolumn{1}{c|}{\textbf{Female}} & \multicolumn{1}{c}{\textbf{Male}} & \multicolumn{1}{c}{\textbf{Female}} \\ \hline
Number of subjects                         & 13                                & 26                                   & 21                                & 18                                   & 24                                & 16                                  \\
Age ($\mathrm{\mu}\pm\mathrm{\sigma}$)                        & 62.8 $\pm$ 10.2                   & 57.7 $\pm$ 11.4                      & 67.4 $\pm$ 12.8                   & 60.5 $\pm$ 8.0                       & 25.3 $\pm$ 4.5                    & 23.1 $\pm$ 3.0                      \\
Age range                                   & 41 -- 81                           & 25 -- 71                              & 49 -- 84                           & 50 -- 74                              & 21 -- 42                           & 20 -- 32                             \\
Time post diagnose (years) ($\mathrm{\mu}\pm\mathrm{\sigma}$) & 8.4 $\pm$ 4.5                     & 13.4 $\pm$ 12.7                      &                                   &                                      &                                   &                                     \\
MDS-UPDRS-III ($\mathrm{\mu}\pm\mathrm{\sigma}$)              & 34.6 $\pm$ 22.1                   & 36.3 $\pm$ 24.2                      &                                   &                                      &                                   &                                     \\
Range of MDS-UPDRS-III                      & 8 -- 82                            & 9 -- 106                              &                                   &                                      &                                   &                                     \\ \hline
\end{tabular}
\end{table*}

\begin{table}[!ht]
\caption{Demographic and clinical information of the participants in the additional validation dataset. \textbf{PD}: Parkinson's disease, \textbf{HC}: healthy controls, \textbf{t}: time post diagnose [years], \textbf{UPDRS-III}: MDS-UPDRS-III.}
\label{tab:test_data}
\centering

\resizebox{1 \linewidth}{!}{
\begin{tabular}{ccc|ccccc}
\hline
 \multicolumn{3}{c}{\textbf{HC}} & \multicolumn{5}{|c}{\textbf{PD}} \\
& \textbf{Gender} & \textbf{Age} & & \textbf{Gender} & \textbf{Age} & \textbf{t} & \textbf{UPDRS-III} \\
\hline

HC 1 & M & 57 & PD 1 & F & 65 & 8   & 24   \\ 
HC 2 & F & 45 & PD 2 & F & 69 & 5   & 26   \\
HC 3 & F & 50 & PD 3 & M & 63 & 6   & 36   \\
HC 4 & F & 57 & PD 4 & M & 76 & 14  & 69   \\
HC 5 & M & 54 & PD 5 & M & 54 & 5   & 43   \\
HC 6 & M & 55 & PD 6 & F & 83 & 6   & 34   \\

\hline
\end{tabular}}
\end{table}

\section{Methods}

\begin{figure*}[!ht]
    \centering
    \includegraphics[width=0.7\linewidth, trim={0cm, 0cm, 0cm, 0cm}, clip]{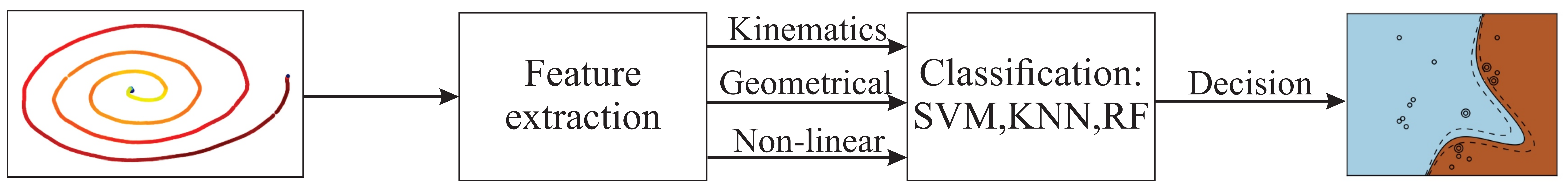}
    \caption{General methodology.}
    \label{fig:methods}
\end{figure*}

The methodology addressed in this work consists of three main stages: feature
extraction, dimensionality reduction, and automatic classification.
This methodology is summarized in Figure~\ref{fig:methods}. Details of each
stage are presented in the following subsections. 

\subsection{Feature Extraction}

Three different feature sets are extracted from the signals: 
features based on kinematic analysis, 
novel features based on the geometrical and spectral analyses of the 
Archimedean spiral, and NLD features.

\subsubsection{kinematic features:}

A total of eight features are extracted: 
(1) absolute/real trajectory of the Archimedean spiral $r(t)$, computed 
as $r(t)=\sqrt{x^2(t)+y^2(t)}$, where $x(t)$ and $y(t)$ are the horizontal and
vertical positions, respectively; 
(2) speed of the trajectory; 
(3) acceleration of the trajectory; 
(4) pressure of the pen; 
(5) first derivative of the pressure; 
(6) second derivative of the pressure; 
(7) distance from the tablet's surface to the pen ($z(t)$); and 
(8) first derivative of $z(t)$.
Six functionals are computed for each feature: mean, standard deviation, maximum value, 
minimum value, skewness, and kurtosis, forming a 48-dimensional feature vector per drawing.

\subsubsection{Geometrical and spectral features}

A novel feature set with geometrical and spectral measures of the 
Archimedean spiral is introduced. 
To create the feature set the spiral's trajectory is modeled as the 
amplitude-modulated signal $\widehat{r}(t)$ defined in
Equation \ref{eq:model_r}. 
\begin{equation}
   \widehat{r}(t)=\left ( a_3t^3+a_2t^2+a_1t+a_0\right)\cdot \sin(2\pi f t)
   \label{eq:model_r}
\end{equation}

\begin{figure*}[!ht]
	\centering
	\includegraphics[width=0.8\linewidth, trim={0cm, 0cm, 0cm, 0cm}, clip]{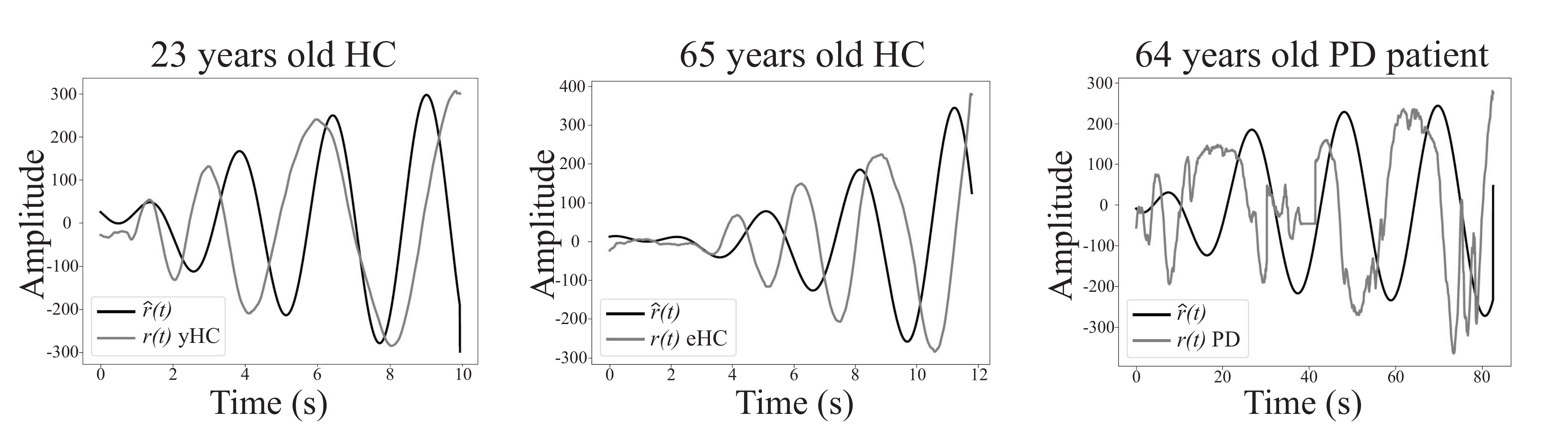}
	\caption{Comparison of the real trajectory of the Archimedean spiral $r(t)$ and the model $\widehat{r}(t)$ for a yHC (left), an eHC (middle) and a PD patient (right).}
	\label{fig:model_tr}
	
\end{figure*}
The real trajectory $r(t)$ is modeled as a sinusoidal signal with increasing amplitude and
frequency $f$. The amplitude values are given by a third-order polynomial which
coefficients ($a_i$) are estimated using a polynomial regression with the maximum 
peaks of the original trajectory. 
A third-order polynomial is chosen because it avoids an oscillatory behavior across the samples. 
Additionally, the third order guarantees a smooth first derivative and a 
continuous second derivative across the trajectory~\cite{de1978practical}.
The frequency $f$ is obtained as the fundamental frequency of the trajectory 
using the Fourier transform.
The model of the trajectory is depicted in Figure~\ref{fig:model_tr} where 
the real and the modeled trajectories can be compared. 
Note that the trajectory for the PD patient is more irregular than those 
observed for the two HC subjects. 
Twelve features are extracted from the modeled trajectory: mean square error 
(MSE) between the real and modeled trajectories, the coefficients 
of the third order polynomial used for the model 
($a_i, i\in \left\lbrace 0,1,2,3 \right\rbrace $), amplitude 
of the first five spectral components of the
trajectory, the slope of the line that links the peaks of the first and third 
spectral components of the trajectory, and the slope of the line that links 
the amplitudes of the third and fifth spectral components 
of the trajectory.

\subsubsection{Non-linear dynamics features}

\begin{figure*}[!ht]
	\centering
	\includegraphics[width=0.8\linewidth, trim={0cm, 0cm, 0cm, 0cm}, clip]{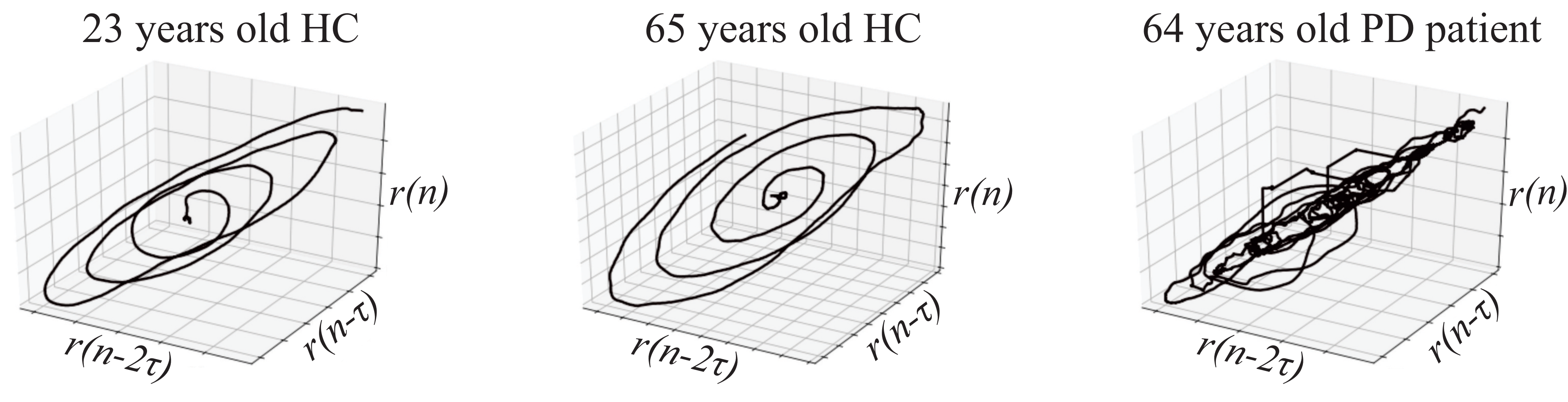}
	\caption{Attractors of the trajectory of the Archimedean spiral for a yHC (left), an eHC (middle) and a PD patient (right).}
	\label{fig:attr}
\end{figure*}

This study considers an exploratory analysis of NLD features to model the handwriting dynamics of PD patients. This approach is motivated by the evidence reported 
	in previous studies~\cite{lopez2016automatic,longstaff1999nonlinear}. Based on these
	reports, we believe that when the disease progresses, the handwriting becomes 
	more distorted and chaotic. Thus, NLD features should be able to reveal specific 
	characteristics in handwriting such that allow us to discriminate between PD patients
	and healthy controls.
The NLD features considered in this study are extracted from the trajectory signal $r(t)$ of 
the Archimedean spiral and from the sentence.
To understand the NLD analysis, the concept of phase space should be 
introduced. It is a multidimensional representation that allows computing 
topological features of a chaotic system. 
For a time series $s(n)$, the phase space can be reconstructed using the 
embedding theorem introduced in~\cite{takens1981detecting}. 
The phase space is defined according to Equation~\ref{eq:att}, 
where $\hat{s}(n)$ is the reconstructed attractor, $n$ is the number of points in 
the time series, and $m$ and $\tau$ are the 
embedding dimension and time delay, respectively. 
The embedding dimension is estimated using the false neighbors 
method~\cite{kennel1992determining}, and the time delay is found as the 
first minimum of the mutual information function.
\begin{equation}
    \hat{s}(n)=\left \{ s(n), s(n-\tau), \cdots, s(n-(m+1)\tau) \right \}
    \label{eq:att}
\end{equation}
Figure~\ref{fig:attr} shows attractors corresponding to trajectories of the 
spirals introduced in Figure~\ref{fig:model_tr}. 
Note that the PD patient exhibits more irregular trajectories in its corresponding
attractor than the HC subjects. 
Once the attractor is created, several features can be 
computed to measure its complexity. In this study a total of seven NLD 
features are extracted from the reconstructed attractors. 

\emph{Approximate entropy (ApEn):}
The ApEn is a regularity statistic to 
measure the average conditional information generated by diverging 
points on trajectories in the attractor. 
A time series containing several repetitive patterns has a relatively small ApEn, 
while a more complex process has a higher ApEn. 
Details of the process to compute the ApEn are described in~\cite{pincus1995approximate}.

\emph{Sample entropy (SampEn):}
The main drawback of ApEn is its dependence on the signal's length due to self
comparison of attractor's points. 
The SampEn is a modification of the ApEn which appears to overcome this drawback. Further details
of the process to compute ApEn can be found in~\cite{richman2000physiological}.

\emph{Approximate and Sample entropies with Gaussian kernels:}
The computation of ApEn and SampEn estimates the regularity of the 
attractor's trajectories by counting neighbor points. 
This process is performed using a Heaviside step function.
Instead of using such a step function, the computation of 
ApEn and SampEn with Gaussian kernels uses the exponential function 
presented in Equation~\ref{eq:GaussianKernel}, where $R$ is a 
tolerance parameter of the distance between near samples in the time-series $s[n]$.
Further details of the computation process can be found in~\cite{xu2005gaussian}.
\begin{equation}
d_G(s[i], s[j])=\exp\left(-\frac{\left|\left|s[i]-s[j]\right|\right|^2}{10R^2}\right)
    \label{eq:GaussianKernel}
\end{equation}
\emph{Correlation Dimension (CD):}
This feature allows to estimate the exact space occupied by the attractor in 
the phase space. 
To estimate CD the correlation sum $C(\varepsilon)$ is defined according to 
Equation~\ref{eq:CS}, where $\Theta$ is the Heaviside step function. 
$C(\varepsilon)$ can be interpreted as the probability to have pairs of  
points in a trajectory of the attractor inside the same sphere 
of radius $\varepsilon$. 
In~\cite{grassberger2004measuring}, the authors demonstrated that 
$C(\varepsilon)$ represents a volume measure, hence $\mathrm{CD}$ 
can be defined by Equation \ref{eq:cd}.

\begin{equation}
C(\varepsilon )=\lim_{n\rightarrow \infty }\frac{1}{n(n-1)}\sum _{i=1}^{n}\sum _{j=i+1}^{n}\Theta (\varepsilon -\left|s[i]-s[j]\right|)
\label{eq:CS}
\end{equation}

\begin{equation}
\mathrm{CD}=\lim_{\varepsilon \rightarrow 0} \frac{\mathrm{log}(C(\varepsilon ))}{\mathrm{log}(\varepsilon)}
\label{eq:cd}
\end{equation}

\emph{Hurst exponent (HE):}
This feature measures the long term dependence of a time series. It is 
defined according to the asymptotic behavior of the re-scaled range of 
a time series as a function of a time interval~\cite{hurst1965long}. 
The estimation process consists of dividing the time series into 
intervals of size $L$ and calculating the average ratio between the range $R$ 
and the standard deviation $\sigma$ of the time series. 
HE is computed as the slope of the curve obtained from Equation~\ref{eq:HE}.

\begin{equation}
L^{\mathrm{HE}}=\frac{R}{\sigma}
\label{eq:HE}
\end{equation}

\emph{Largest Lyapunov exponent (LLE):}
This  feature  represents the average divergence rate of neighbor 
trajectories in the phase space. 
Its estimation process follows the algorithm in \cite{rosenstein1993practical}. 
After the reconstruction of the phase space, the nearest neighbor of every 
point in the trajectory is estimated. The LLE is estimated as the average 
separation rate of those neighbors in the phase space.

\emph{Lempel-Ziv complexity (LZC):}
This feature measures the degree of disorder of spatio-temporal 
patterns in a time series~\cite{lempel1976complexity}. 
In the computation process the signal is transformed into binary 
sequences according to the difference between consecutive samples, and 
the LZC reflects the rate of new patterns in
the sequence. It ranges from 0 (deterministic sequence) to 1 (random sequence).
Further details of the computation process can be found in~\cite{Kaspar-1987,Travieso-2017}.


\subsection{Classification}

Three different classifiers are used in this study: 
(1) K-nearest neighbors, (2) SVM with a Gaussian kernel, and 
(3) random forest (RF). 
The three aforementioned classifiers were trained and tested 
following a leave one out cross-validation strategy. 
This procedure is repeated for each sample to assure that 
all data are tested. The parameters of the classifiers are optimized
in a grid-search. For the KNN the possible number of neighbors was 
$K\in\{3,5,7\}$. 
For the SVM the parameters $C$ and $\gamma$ were optimized up to 
powers of ten where $C\in\{0.0001, 0.001,\cdots,1000\}$ and 
$\gamma\in\{0.0001, 0.001,\cdots,1000\}$. 
Finally, for the RF the number of trees and their maximum depth were 
$N\in\{5,10,15,20,50\}$ and $D\in\{1,2,5,10\}$, respectively. 
The performance of the classifiers was evaluated considering several statistics 
including the F-score, accuracy, sensitivity, and specificity.

\section{Experiments and results}

A total of five experiments are performed in this study. Four of them comprise bi-class
classification experiments: PD vs. yHC; PD vs. eHC; relevance analysis; and 
classification of PD vs. HC considering the separate validation sets described in Table~\ref{tab:test_data}. The fifth experiment is a multi-class classification task where the goal is to
differentiate among four different stages of the disease. Details of all these 
experiments are described in the next subjections.

\subsection{Classification of PD vs. yHC}
Table~\ref{tab:classpdhcy} shows the results when PD patients vs. yHC subjects 
are classified using the aforementioned classifiers using different feature
We performed a Kruskal-Wallis test to asses whether a significant difference existed between the features computed from the PD patients and yHC subjects. The null hypothesis of both feature sets coming from the same distribution was rejected in all cases ($p\ll0.05$).
The results obtained after the relevance analysis with PCA 
(see section~\ref{sec:pca}) 
are also included in this table. 
The best results are obtained with the features computed 
from the Archimedean spiral ($F1=0.94$). 
Note that in general the SVM and the RF classifiers provide the highest 
accuracies. Note also that there is a low variation of the optimal 
hyper-parameters along the feature sets and tasks, which gives an idea about the 
stability and robustness of the proposed approach. 

\begin{table*}[!th]
\centering
\caption{Classification of PD patients vs. yHC subjects using SVM, KNN, and RF classifiers. \textbf{Kinem: }kinematic features, \textbf{Geom: }Geometrical features, \textbf{NLD: }NLD features, \textbf{Acc: }Accuracy, \textbf{Spec. }Specificity, \textbf{Sens. }Sensitivity, \textbf{F1: }F1 score}
\label{tab:classpdhcy}
\resizebox{\linewidth}{!}{
\begin{tabular}{ll|llcccc|ccccc|lccccc}
\hline
\multicolumn{1}{c}{\textbf{}} & \multicolumn{1}{c|}{\textbf{}}   & \multicolumn{6}{c|}{\textbf{SVM}}          & \multicolumn{5}{c|}{\textbf{KNN}} & \multicolumn{6}{c}{\textbf{RF}}           \\ 
\multicolumn{1}{c}{Task}      & \multicolumn{1}{c|}{Feature set} & C  & $\gamma$     & Acc (\%) & Spec  & Sens  & F1   & K & Acc (\%) & Spec  & Sens  & F1   & N  & D & Acc (\%) & Spec  & Sens  & F1   \\ \hline
Spiral                        & Kinem.                           & 10 & 0.01  & \bf 94.0     & 0.94 & 0.94 & 0.94 & 3 & 86.0     & 0.89 & 0.86 & 0.86 & 20 & 5     & \bf 92.4     & 0.92 & 0.92 & 0.92 \\
Spiral                        & Geom.                            & 1  & 0.1   & 78.5     & 0.79 & 0.78 & 0.78 & 5 & 72.2     & 0.73 & 0.72 & 0.72 & 5  & 2     & 77.2     & 0.78 & 0.77 & 0.77 \\
Spiral                        & Kinem.+Geom.                     & 10 & 0.01  & 93.7     & 0.94 & 0.94 & 0.94 & 5 & 86.1     & 0.88 & 0.86 & 0.86 & 5  & 1     & 83.5     & 0.84 & 0.84 & 0.84 \\
Spiral                        & NLD                              & 1  & 0.1   & 77.2     & 0.77 & 0.77 & 0.77 & 5 & 69.6     & 0.71 & 0.70 & 0.69 & 5  & 1     & 67.1     & 0.67 & 0.67 & 0.67 \\
Spiral                        & Kinem.+Geom.+NLD                 & 10 & 0.01  & 91.1     & 0.91 & 0.91 & 0.91 & 5 & 87.3     & 0.90 & 0.87 & 0.87 & 15 & 1     & 88.6     & 0.89 & 0.89 & 0.89 \\

Spiral                        & PCA Kinem.+Geom.+NLD                 & 1 & 0.01  & 93.7     & 0.94 & 0.94 & 0.94 & 5 &  \bf 88.6     & 0.90 & 0.89 & 0.89 & 10 & 5     & 91.6    & 0.92 & 0.92 & 0.92 \\

Sentence                      & Kinem.                           & 1  & 0.01  & 92.0     & 0.92 & 0.92 & 0.92 & 3 & 86.7     & 0.88 & 0.87 & 0.86 & 50 & 5     & 90.7     & 0.91 & 0.91 & 0.91 \\
Sentence                      & NLD                              & 10 & 0.1   & 81.3     & 0.81 & 0.81 & 0.81 & 5 & 77.3     & 0.78 & 0.77 & 0.77 & 20 & 1     & 74.7     & 0.75 & 0.75 & 0.74 \\
Sentence                      & Kinem.+NLD                       & 10 & 0.001 & 93.3     & 0.93 & 0.93 & 0.93 & 3 & 85.3     & 0.86 & 0.85 & 0.85 & 50 & 5     & 89.3     & 0.89 & 0.89 & 0.89 \\ \hline
\end{tabular}}
\end{table*}

\subsection{Classification of PD vs. eHC}

The Kruskal-Wallis test to assess the significant difference between the features computed from the PD patients and eHC subjects reflect that both feature sets do not come from the same distribution ($p\ll0.05$).
The results of the classification between PD patients and eHC subjects 
are shown in Table~\ref{tab:classpdhce}. 
This table also includes the results 
obtained after the relevance analysis with PCA (see section~\ref{sec:pca}). 
Note that the results are slightly lower than those obtained in the 
discrimination of PD patients vs. yHC subjects.
Although such a negative impact, high accuracies were also 
obtained when classifying PD patients vs. elderly HC subjects (F1 score of up to 0.87). The best results are obtained with the kinematic and 
kinematic + Geometrical feature sets. 
Similar to the previous case, the best classifiers in this experiment 
were the SVM and the RF. 
Note also that the dispersion of the hyper-parameters in this case 
is much larger than in the previous experiment, which gives count of the
complexity of the classification problem. Note that  
the values of $\gamma$ are, smaller than those in the previous experiment, 
indicating that the models 
are constrained and the set of support vectors should include 
a large number of training samples.

\begin{table*}[!th]
\centering
\caption{Classification of PD patients vs. eHC subjects using SVM, KNN, and RF classifiers. \textbf{Kinem: }kinematic features, \textbf{Geom: }Geometrical features, \textbf{NLD: }NLD features, \textbf{Acc: }Accuracy, \textbf{Spec. }Specificity, \textbf{Sens. }Sensitivity, \textbf{F1: }F1 score}
\label{tab:classpdhce}
\resizebox{\linewidth}{!}{
\begin{tabular}{ll|llcccc|ccccc|llcccc}
\hline
\multicolumn{1}{c}{\textbf{}} & \multicolumn{1}{c|}{\textbf{}}   & \multicolumn{6}{c|}{\textbf{SVM}}          & \multicolumn{5}{c|}{\textbf{KNN}} & \multicolumn{6}{c}{\textbf{RF}}           \\ 
\multicolumn{1}{c}{Task}      & \multicolumn{1}{c|}{Feature set} & C  & $\gamma$     & Acc (\%) & Spec  & Sens  & F1   & K & Acc (\%) & Spec  & Sens  & F1   & N  & D & Acc (\%) & Spec  & Sens  & F1   \\ \hline
Spiral                        & Kinem.                           & 100    & 0.001  & 87.0     & 0.87 & 0.87 & 0.87 & 3 & \bf 86.7     & 0.87 & 0.86 & 0.86 & 10 & 1     & \bf 85.3     & 0.85 & 0.85 & 0.85 \\
Spiral                        & Geom.                            & 100    & 0.0001 & 30.7     & 0.74 & 0.69 & 0.72 & 3 & 53.3     & 0.53 & 0.53 & 0.53 & 5  & 1     & 57.3     & 0.58 & 0.58 & 0.59 \\
Spiral                        & Kinem.+Geom.                     & 10     & 0.001  & 86.7     & 0.89 & 0.87 & 0.87 & 5 & 81.3     & 0.84 & 0.81 & 0.81 & 15 & 1     & 84.0     & 0.84 & 0.84 & 0.84 \\
Spiral                        & NLD                              & 0.0001 & 0.0001 & 52.0     & 0.73 & 0.52 & 0.64 & 7 & 58.7     & 0.59 & 0.59 & 0.59 & 5  & 1     & 56.0     & 0.58 & 0.56 & 0.58 \\
Spiral                        & Kinem.+Geom.+NLD                 & 100    & 0.0001 & 86.7     & 0.89 & 0.87 & 0.87 & 5 & 80.0     & 0.81 & 0.80  & 0.80  & 10 & 1     & 84.0     & 0.84 & 0.84 & 0.84 \\

Spiral                        & PCA Kinem.+Geom.+NLD                 & 10 & 0.001  & \bf 89.3     & 0.89 & 0.89 & 0.89 & 3 & 81.3     & 0.81 & 0.81 & 0.81 & 50 & 5     & 81.0    & 0.81 & 0.81 & 0.81 \\

Sentence                      & Kinem.                           & 1      & 0.01   & 70.4     & 0.70 & 0.70 & 0.70 & 7 & 73.2     & 0.73 & 0.73 & 0.73 & 15 & 10    & 78.9     & 0.79 & 0.79 & 0.79 \\
Sentence                      & NLD                              & 100    & 0.01   & 62.0     & 0.62 & 0.62 & 0.62 & 5 & 63.4     & 0.63 & 0.63 & 0.63 & 50 & 1     & 60.6     & 0.61 & 0.61 & 0.61 \\
Sentence                      & Kinem.+NLD                       & 0.001  & 0.001  & 53.5     & 0.75 & 0.53 & 0.68 & 7 & 78.9     & 0.80  & 0.79 & 0.79 & 20 & 5     & 73.2     & 0.73 & 0.73 & 0.73 \\ \hline

\end{tabular}}
\end{table*}

\begin{figure}[!ht]
	\centering
	\includegraphics[width=1\linewidth, trim={0cm, 0cm, 0cm, 0cm}, clip]{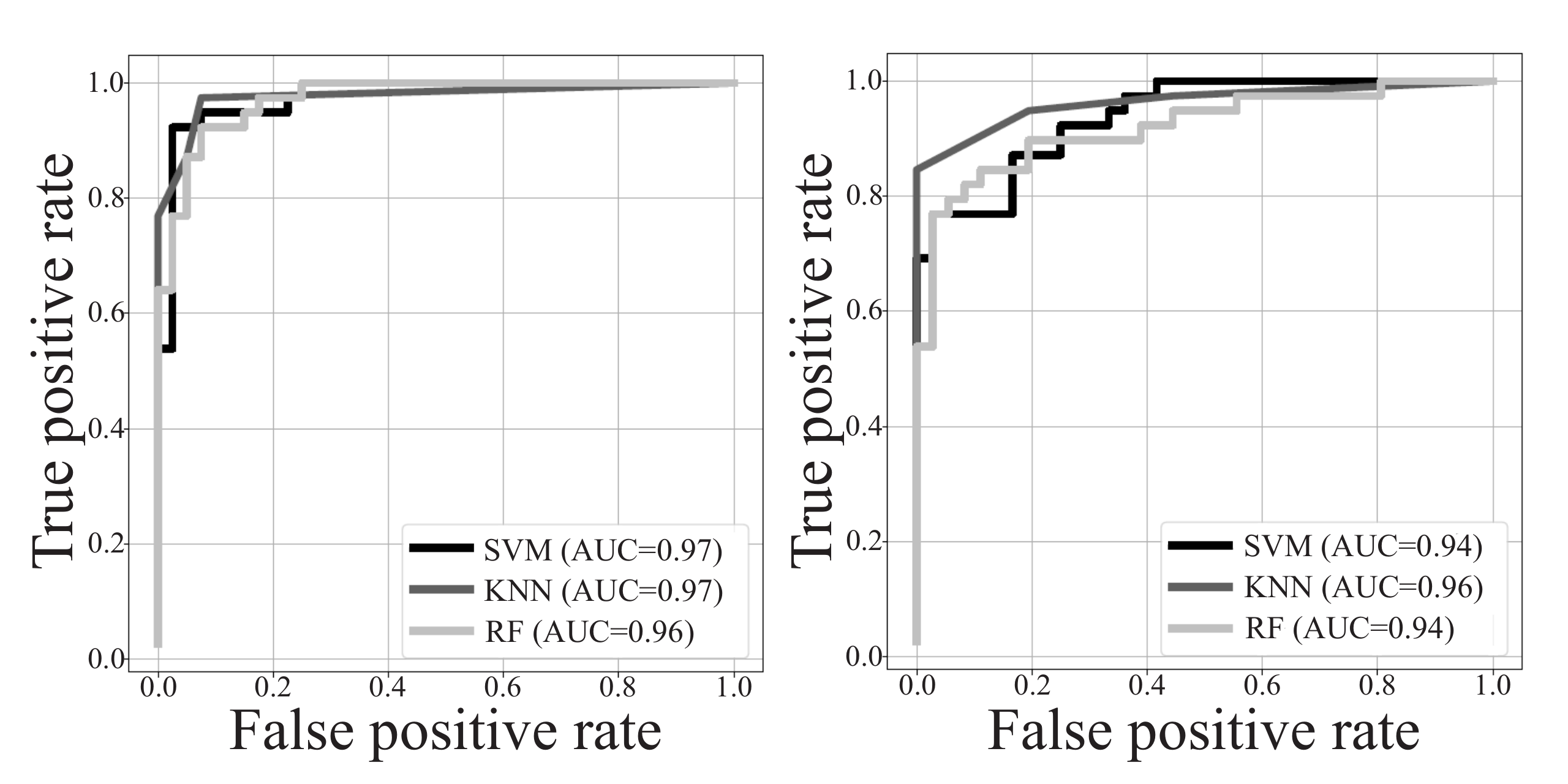} 
	\caption{ROC curves for the classification of PD patients vs. young HC subjects (left) and for PD patients vs. elderly HC subjects (right).}
	\label{fig:rocs}
\end{figure}

\begin{figure}[!ht]
	\centering
	\includegraphics[width=1\linewidth, trim={0cm, 0cm, 0cm, 0cm}, clip]{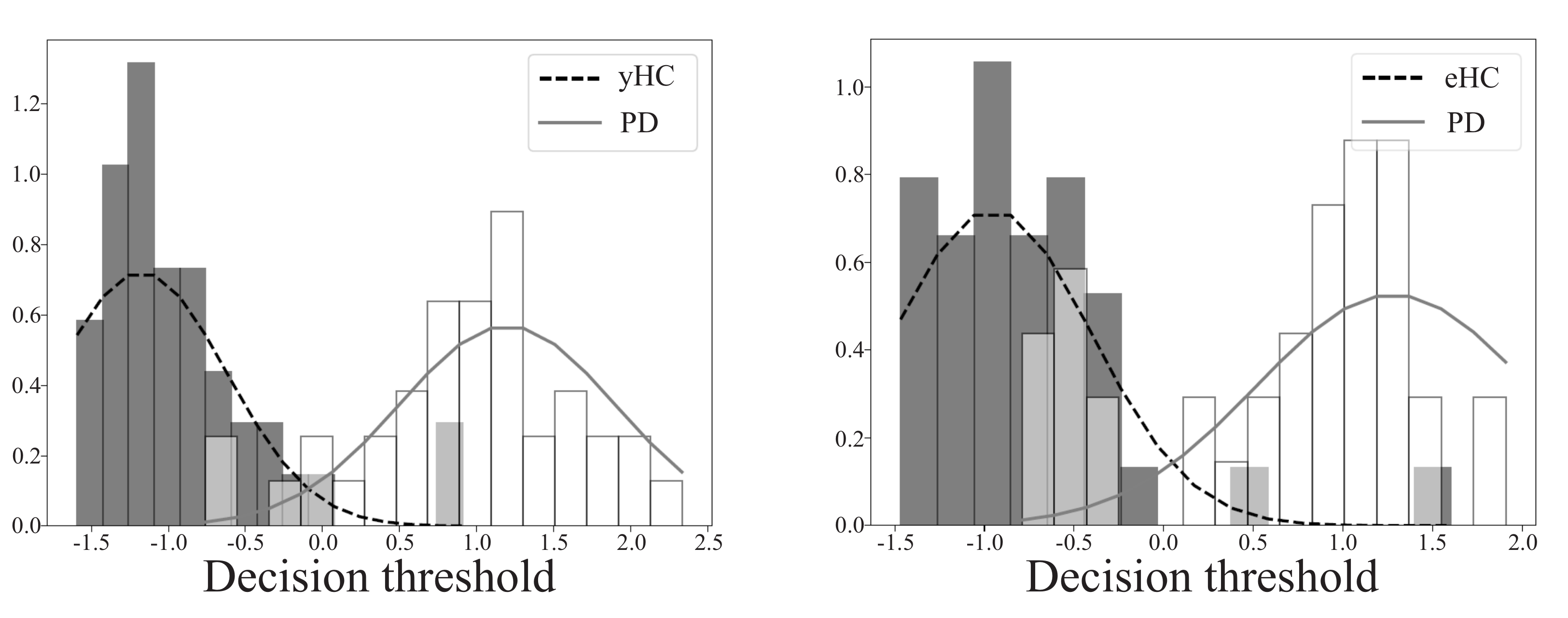} 
	\caption{Histograms and the corresponding probability density distributions of the scores obtained from PD patients and yHC subjects (left), and for PD patients and eHC subjects (right).}
	\label{fig:hist}
\end{figure}

Figure \ref{fig:rocs} shows the Receiver Operating Characteristic (ROC) curves 
obtained from the previously described classification experiments. These curves allow
the comparison among results obtained in the 
classification of PD patients and the two groups of HC subjects. 
Only results from kinematic and geometrical features extracted from the 
Archimedean spiral are included. 
It can be observed that the three classifiers provide similar results 
in both experiments. The results when classifying PD vs. eHC 
confirm the negative impact of aging in the classification process.
Histograms and the corresponding fitted probability density distribution are shown 
in Figure~\ref{fig:hist}. The figure illustrates the distribution of 
the SVM scores, i.e., the distance of data samples to the separating hyperplane. The statistical significance of the results was assessed with a Welch t-test to evaluate the difference in the scores obtained for PD patients and HC subjects. The null hypothesis was rejected for PD vs. yHC ($t=-9.043$, $p\ll0.005$), and for PD vs. eHC ($t=-12.817$, $p\ll0.005$).

\subsection{Relevance analysis}
\label{sec:pca}

Apart from the typical classification between groups of subjects, we want to
perform a relevance analysis such that allows the identification of the most
relevant/discriminant features. This analysis help in finding features that
could be potentially used in clinical practice not only to make informed decisions
but to provide additional information to the clinician about the health state
of a patient. In this paper the process is based on the Principal Component 
Analysis (PCA) algorithm.
The original feature matrix $X\in\mathbb{R}^{m\times p}$ is reduced to a matrix $X_r\in\mathbb{R}^{m\times q},\, (q<p)$ according to a relevance vector $\mathbf{\rho}=[\rho_1,\rho_2,\cdots,\rho_q,\cdots,\rho_p]$ computed as 
$\mathbf{\rho}=\sum_{j=1}^p \left | \lambda_j \mathbf{v}_j \right |$, 
where $\lambda_j$ and $\mathbf{v}_j$ are the eigenvalues and the eigenvectors of the original feature matrix~\cite{daza2009dynamic}. 
The values of $\mathbf{\rho}$ for each feature are related with the 
contribution of the feature to each principal component. 
The original features with higher $\rho$ are the most correlated 
with the principal components and are included to form the reduced feature space.
This approach has been successfully used in previous studies and its main advantage
over other existing feature selection methods is its low computational 
cost~\cite{Orozco-2011, Arias2017}.
In this work the relevance analysis is performed upon the feature space
formed with kinematic, geometrical and NLD features extracted from 
	the Archimedian spiral. 90\% of the total 
variance is kept in the reduced feature space which is formed with a total 
of 18 features for the case of PD vs. yHC, 19 features for PC vs. eHC, and 
16 features for yHC vs. eHC.
Table \ref{tab:pca1} indicates the top 10 of the selected features 
on each case. 

\begin{table}[!ht]
    \caption{Top 10 of selected features from the Archimedian spiral. $\Delta$ and $\Delta\Delta$ indicate the first and second 
    	derivative, respectively.}
    \centering
    \begin{tabular}{l|l}
    \hline
    \textbf{PD vs. yHC} & \textbf{PD vs. eHC} \\ \hline
    Min. $\Delta\Delta$pressure & $a_1$ of $\,\widehat{r}(t)$ \\
    Average speed & Skewness of acceleration \\
    Average pressure & Kurtosis of acceleration \\
    Kurtosis of speed & MSE between $r(t)$ and $\widehat{r}(t)$ \\
    Min. speed & Average acceleration  \\
    Skewness of $r(t)$ & Std. of $\Delta z(t)$ \\
    Skewness of $\Delta\Delta$pressure & 2nd spectral component of $r(t)$ \\
    Max. speed & Max. acceleration \\
    Min. $\Delta z(t)$ & Min. acceleration \\
    Std. of $\Delta z(t)$ & LZC \\
    \hline
    \end{tabular}
    \label{tab:pca1}
\end{table}

The results indicate that the best features to classify PD patients and yHC
subjects are those based on the kinematic analysis. 
When considering the case of PD patients vs. eHC, some 
of the geometrical and spectral features appear in the top ten of the 
selected features. Particularly, the first coefficient of the third-order
polynomial ($a_1$) used to model the amplitude of the trajectory 
$\widehat{r}(t)$ appears in the first place of the top ten. 
Finally, note that the features selected in the case of eHC vs. yHC 
are mostly different compared to those selected in the previous cases.

\begin{figure*}[!ht]
	\centering
	\setlength\tabcolsep{1pt} 
	\begin{tabular}{cc}
		A & B\\
		\includegraphics[width=0.4\linewidth]{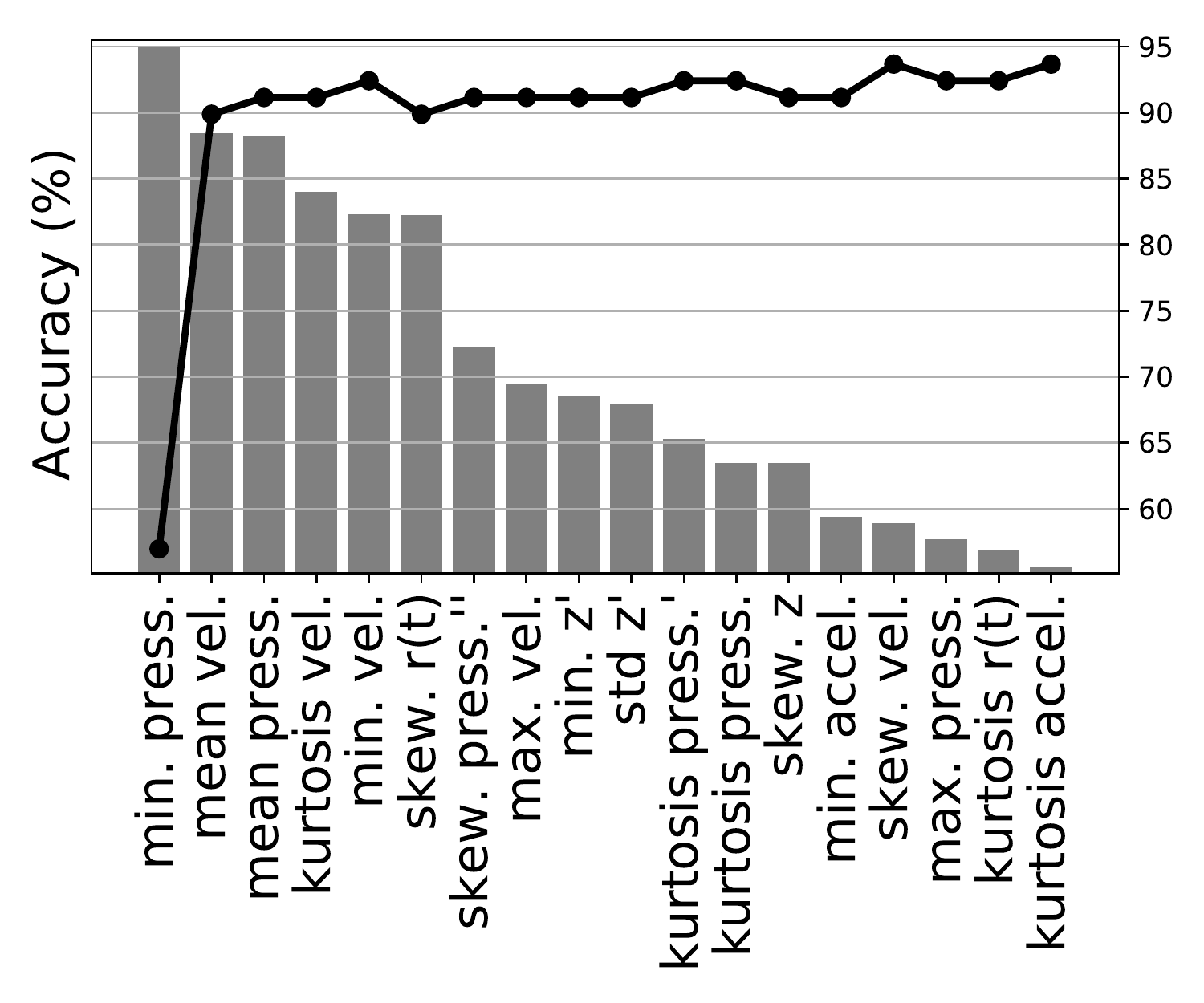}&
		\includegraphics[width=0.4\linewidth]{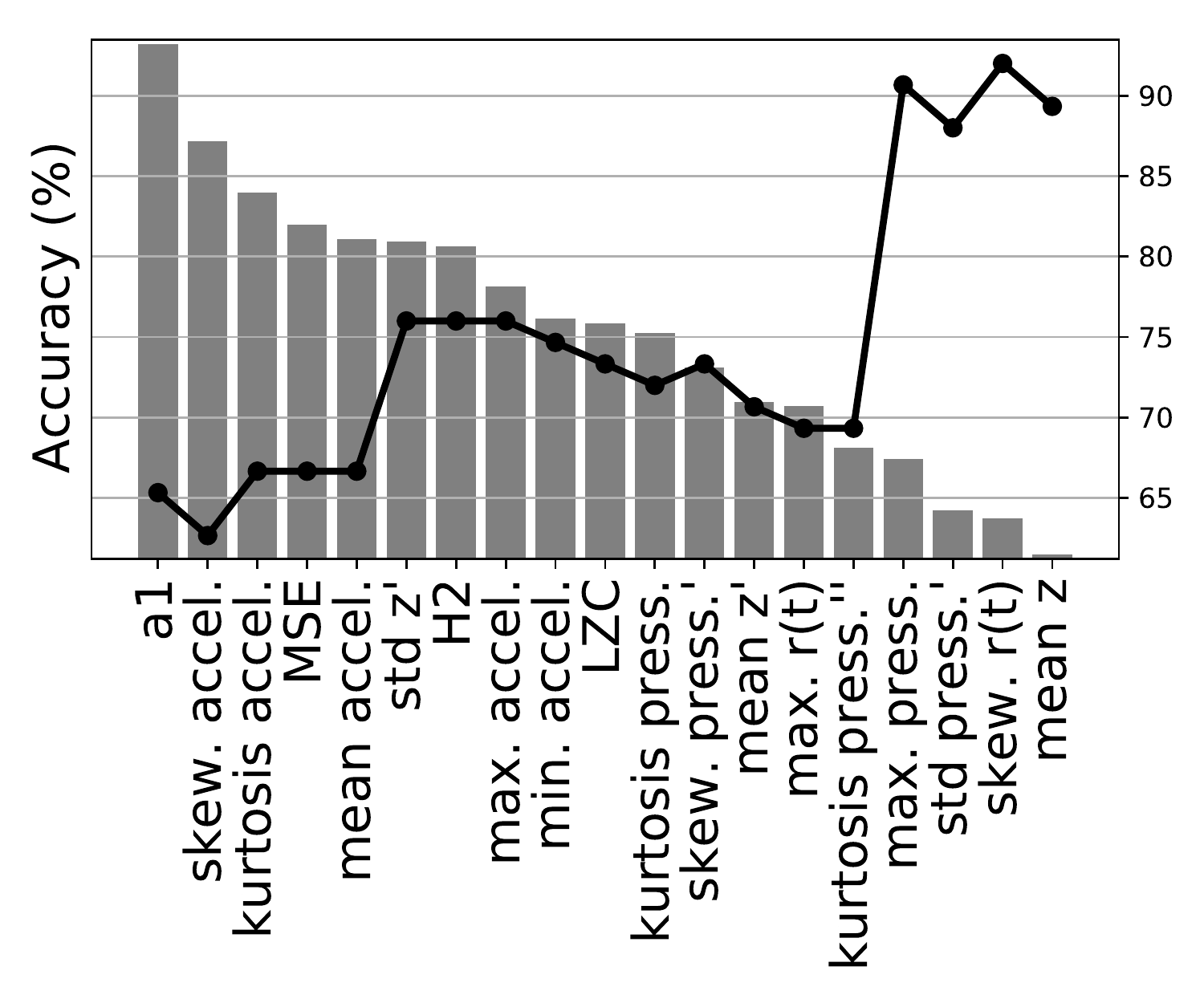}
	\end{tabular}
	\caption{Classification of PD vs. yHC subjects (left) and PD vs. eHC subjects (right) when the selected features are sequentially added.}
	\label{fig:pcafeat}
\end{figure*}

The selected features were used to perform two classification experiments:
PD vs. yHC and PD vs. eHC. Results are indicated in Tables~\ref{tab:classpdhcy}
and \ref{tab:classpdhce}, respectively.
The aim is to evaluate the contribution of each selected feature in 
these classification tasks. 
Each feature was sequentially added according to the order given by 
the relevance factor $\rho$. The classification step is performed using 
a SVM. 
The results are shown in Figure \ref{fig:pcafeat}A for PD vs. yHC, and 
in Figure \ref{fig:pcafeat}B for PD vs. eHC. 
The bars indicate the relevance factor of each feature and the 
black lines indicate the obtained incremental accuracy.
Note that when classifying PD vs. yHC about 90\% of accuracy is 
obtained already with the first two relevant features, indicating that
the problem is relatively simple and it can be solved with a low dimensional
(less complex) feature space.
Conversely, when classifying PD vs. eHC at least the first 16 features 
are required to reach accuracies of around 90\%. This fact
confirms the increased complexity of that problem due to the impact 
of aging in the handwriting process. 
We performed additional experiments (not reported here) using only those 
features that provide an incremental improvement in the accuracy, 
however, the results were not satisfactory, indicating that all of 
the selected features are relevant for the classification process.  
\subsection{Classification using a separate validation set}

This experiment is performed with the aim of evaluating
the proposed approach in a more realistic scenario. 
Given that this set of additional subjects was recorded
in sessions different than those of the other groups, it is possible to say
that this additional validation set represents a group of subjects who arrived in
the clinics and performed a handwriting test to decide whether to continue with 
further screenings to validate their neurological condition.
The results are shown in Table~\ref{tab:classpdhctest}. 
Note that the parameters of the classifiers are 
	the same as those used in 
Table~\ref{tab:classpdhce}, which means that no further optimization was 
performed over the system, i.e., the patients in the separated validation
set never participated in the configuration/optimization of the classifiers.
The accuracy obtained with the KNN and RF classifiers are 
above 83\% in several cases 
with different feature sets extracted from the Arquimedian spiral.
These results are comparable to those obtained in the PD vs. eHC experiments, 
which confirms the generalization capability of the proposed approach when using
the KNN and RF classifiers. 
The results with the SVM are not as high as with the other classifiers 
perhaps because the optimization of the two meta-parameters needs a 
fine tunning which was not performed in our experiments to avoid biased results 
due to over-fitting. Further experiments will be performed in the near future with data 
from more subjects in order to improve the stability and robustness of this classifier.

\begin{table*}[!ht]
\centering
\caption{Classification of PD patients vs. HC subjects using the separate validation set. \textbf{Kinem:} kinematic features, \textbf{Geom:} Geometrical features, \textbf{NLD:} NLD features, \textbf{Acc:} Accuracy, \textbf{Spec.} Specificity, \textbf{Sens.} Sensitivity, \textbf{F1:} F1 score}
\label{tab:classpdhctest}
\resizebox{\linewidth}{!}{
\begin{tabular}{ll|llcccc|ccccc|lccccc}
\hline
\multicolumn{1}{c}{}     & \multicolumn{1}{c|}{}            & \multicolumn{6}{c|}{SVM}                          & \multicolumn{5}{c|}{KNN}          & \multicolumn{6}{c}{RF}                \\
\multicolumn{1}{c}{Task} & \multicolumn{1}{c|}{Feature set} & C      & $\gamma$ & Acc (\%) & Spec & Sens  & F1   & K & Acc (\%) & Spec & Sens  & F1   & N  & D & Acc (\%) & Spec & Sens  & F1   \\ \hline
Spiral                   & Kinem.                           &  100      & 0.001         & 50.0  &       0.50 & 0.50     & 0.50         & 3   & \bf 83.3         & 0.83      & 0.83     & 0.83      &  10  & 1  &    66.6      & 0.69     & 0.66     & 0.66     \\
Spiral                   & Geom.                            &  100      &    0.0001      &    58.3      & 0.23     & 0.41     & 0.29     & 3  & 58.3         &    0.59  & 0.58     & 0.58     & 5   &1   &    58.3      & 0.61     & 0.58     & 0.55     \\
Spiral                   & Kinem.+Geom.                     & 10     & 0.001        & 66.6     & 0.69 & 0.67 & 0.66 & 5 & 81.3     & 0.84 & 0.81 & 0.81 & 15 & 1 & \bf 83.3     & 0.93 & 0.92 & 0.92 \\
Spiral                   & NLD                              & 0.0001 & 0.0001   & 50.0     & 0.75 & 0.50 & 0.66 & 7 & 58.4     & 0.59 & 0.58 & 0.59 & 5  & 1 & 64.3     & 0.65 & 0.65 & 0.65 \\
Spiral                   & Kinem.+Geom.+NLD                 & 100    & 0.0001   & 58.3    & 0.58 & 0.58 & 0.58 & 5 & \bf 83.3     & 0.83 & 0.83 & 0.83 & 10 & 1 & 72.0     & 0.71 & 0.71 & 0.71 \\

Spiral                   & PCA Kinem.+Geom.+NLD                 & 10    & 0.001   & 66.6    & 0.66 & 0.66 & 0.66 & 3 & 66.6     & 0.66 & 0.66 & 0.66 & 50 & 5 & 50.0     & 0.50 & 0.50 & 0.52 \\

Sentence                 & Kinem.                           & 1      & 0.01     & \bf 75.0     & 0.76 & 0.75 & 0.75 & 7 & 58.3     & 0.61 & 0,42 & 0.72 & 15 & 10 & 58.4     & 0.50 & 0.50 & 0.50 \\
Sentence                 & NLD                              & 100    & 0.01     & 58.3     & 0.59 & 0.59 & 0.59 & 5 & 59.0     & 0.59 & 0.59 & 0.61 & 50 & 1 & 51.2     & 0.51 & 0.51 & 0.51 \\
Sentence                 & Kinem.+NLD                       & 0.001 & 0.001     & 50.0     & 0.75 & 0.50 & 0.66 & 7 & 50.0     & 0.50 & 0.50 & 0.67 & 20 & 5 & 51.3     & 0.51 & 0.51 & 0.51 \\ \hline
\end{tabular}}
\end{table*}
\subsection{Classification of PD patients in different stages of the disease}
Four-class classification experiments were performed considering 
four groups: (1) HC subjects; (2) PD patients with 
MDS-UPDRS-III scores below 20 (initial stage--PD1); 
(3) PD patients with MDS-UPDRS-III scores between 20 and 40 
(intermediate stage--PD2); and (4) PD patients with MDS-UPDRS-III scores 
above 40 (advance stage--PD3). 
Classification was performed with a multi-class SVM 
following a one vs. all strategy.
Confusion matrices with the results are reported in Table~\ref{tab:multiclass}.
Results are presented in terms of accuracy (Acc), F1 score (F1), and the Cohen-Kappa index ($\kappa$).

\begin{table*}[!ht]
\centering
\scriptsize
\caption{Confusion matrices with results of classifying HC subjects and 
	PD patients in different stages of the disease. {PD1}: patients with
	 MDS-UPDRS-III scores between 0 and 20. {PD2}: patients with  MDS-UPDRS-III scores between 21 and 40. {PD3}: patients with  MDS-UPDRS-III scores above 40.
	 Spiral PCA indicates the results obtained with the set of 19 features
	 that results after the feature selection process when classifying PD vs. eHC
	 (see Figure~\ref{fig:pcafeat}).
	  The results are expressed in (\%). F1: F1 score. $\kappa$: Cohen-kappa index.}
\label{tab:multiclass}
\resizebox{\linewidth}{!}{
\begin{tabular}{l|llll|llll|llll|llll}
\hline
             & \multicolumn{4}{c|}{Spiral Kinem.}                    & \multicolumn{4}{c|}{Spiral Kinem.+Geom.}              & \multicolumn{4}{c|}{Spiral Kinem+Geom+NLD}            & \multicolumn{4}{c}{Spiral PCA}                       \\
             & \multicolumn{4}{c|}{{Acc=61.3, F1=0.58, $\kappa$=0.40}} & \multicolumn{4}{c|}{{Acc=64.0, F1=0.62, $\kappa$=0.44}} & \multicolumn{4}{c|}{{Acc=64.0, F1=0.62, $\kappa$=0.44}} & \multicolumn{4}{c}{{Acc=57.3, F1=0.55, $\kappa$=0.36}} \\
             & {HC}   & {PD1}   & {PD2}  & {PD3}  & {HC}   & {PD1}   & {PD2}  & {PD3}  & {HC}   & {PD1}   & {PD2}  & {PD3}  & {HC}   & {PD1}   & {PD2}  & {PD3}  \\ \hline
{HC}  & {97.2}          & 0.0            & 2.8           & 0.0           & {97.2}          & 0.0            & 2.8           & 0.0           & 97.2          & 0.0            & 2.8           & 0.0           & {91.7}          & 2.8            & 5.5           & 0.0           \\
{PD1} & 15.4          & {30.8}           & 15.4          & {38.5}          & 15.4          & {38.5 }          & 7.7           & 38.5          & 15.4          & 38.5           & 15.4          & 30.8          & 15.4          & {38.5 }          & 15.4          & 30.8          \\
{PD2} & 21.4          & 28.6           & {35.7}          & 14.3          & 21.4          & 14.3           & {28.6}          & 35.7          & 21.4          & 21.4           & {28.6}          & 28.6          & 28.6          & 28.6           & {28.6}          & 14.3          \\
{PD3} & 33.3          & 33.3           & 16.7          & {16.7}          & 33.3          & 16.7           & 16.7          & {33.3}          & 33.3          & 16.7           & 16.7          & {33.3}          & 0.0           & 50.0           & 41.7          & {8.3}          \\ \hline
\end{tabular}}
\end{table*}

In this case only drawings of the Archimedean spiral were considered and 
only the feature sets that exhibited the best results in the experiments
with the separate validation set were extracted: kinematic; 
kinematic + geometrical; and kinematic + geometrical + NLD 
(see Table~\ref{tab:classpdhctest}).
Additionally, we explored the suitability of the feature set selected
with PCA which was previously used to 
classify between HC and eHC subjects (Section~\ref{sec:pca}).
The results indicate that HC subjects can be accurately classified,
while the patients in advance stage (PD3) are the most difficult to be
discriminated. 
The results with kinematic + geometrical features are similar to those
obtained with kinematic + geometrical + NLD features, which likely indicates 
that the nonlinear features are not complimentary to 
the kinematic and geometrical ones, at least to discriminate different stages of
PD. 
The confusion matrix also indicates that dimensionality reduction does
not have positive impact in the results, conversely increases the
false positive rate making the system to confuse HC subjects with 
patients in initial (PD1) or intermediate (PD2) stages of the disease.

\section{Discussion}
Kinematic features seem to be the most suitable to perform the automatic discrimination
between PD patients and HC subjects (young or elderly). The combination of the
three feature sets in the same space also exhibited good classification results.
Besides the binary classification, a relevance analysis was performed.
According to the results, the most discriminative features are geometrical and kinematic. 
Most of the selected features as relevant are related to speed, pressure, and acceleration of the 
strokes, which is related to the deficits exhibited by patients when performing motor activities
like handwriting.
Patients in different stages of the disease were also classified and $\kappa$ indexes between $0.36$
and $0.44$ were obtained. The results indicate that, at least in the experiments performed here, 
the nonlinear features do not contribute to improve the classification of different
stages of the disease.
An additional experiment with a separate validation set with PD and HC subjects was performed. 
This validation set was built during recording sessions different than those performed
to collect the signals considered in the other experiments. Thus this additional experiment allows
the evaluation of the proposed approach in real conditions. 
The results show that it is possible to discriminate between PD and HC subjects (in the separate validation set)
with accuracies of up to 83.3\%. To the best of our knowledge, this is the first work that considers experiments
with separate and independent validation samples. 
Further research should be performed by combining the proposed model with novel approaches based on deep learning methods~\cite{vasquez2018multimodal,Pereira2018}. In addition, longitudinal studies should be performed to understand and track the progress of the disease in the PD patients through time.

\section{Conclusion}

This paper explored and evaluated the suitability of three different feature sets 
(kinematic, geometrical and NLD) to model handwriting impairments exhibited by PD patients. We provide an accurate method to classify between Parkinson's patients and healthy subjects. The model was validated in a different dataset, and high accuracies were obtained (83.3\%). The classification of Parkinson's patients in several stages of the disease is promising. We are aware of the fact that more research and a larger sample of patients is necessary to lead to 
more conclusive results; however, we think that the results presented here are a step forward to the development of non-intrusive methods, useful in clinical practice, to diagnose and monitor Parkinson's patients.

Future studies will include the comparison of the proposed approach with recent studies based on deep learning strategies, which have shown to be also accurate to model the handwriting deficits of PD patients.

\section*{Acknowledgments}

The authors thank to the patients of Fundalianza Parkinson Colombia for their invaluable support of this study.
This work was financed by CODI from UdeA grants 
PRG2015--7683 and PRV16-2-01, and from the European Union’s Horizon 2020 research and innovation 
programme under the Marie Sklodowska-Curie Grant Agreement No. 766287. 

\emph{Ethical Approval and informed consent:} 
Informed consent was obtained from all participants of the
study. All procedures were in accordance with the ethical standards 
of the institutional research committee and with the 1964 
Helsinki declaration and its later amendments. 
The procedures were approved by the Ethical Research
Committee of the University of Antioquia.

\section*{References}
\bibliographystyle{elsarticle-num}
\bibliography{test.bib}

\end{document}